\documentclass[letterpaper]{article} 
\usepackage{graphicx}
\usepackage{subcaption} 
\usepackage{amsmath}
\usepackage[table]{xcolor}
\usepackage{makecell}
\usepackage{pifont}
\usepackage{booktabs}
\usepackage{multirow}
\usepackage{aaai25}  
\usepackage{times}  
\usepackage{helvet}  
\usepackage{courier}  
\usepackage[hyphens]{url}  
\usepackage{graphicx} 
\usepackage{color}
\usepackage{natbib}  
\usepackage{caption} 
\usepackage{algorithm}
\usepackage{algorithmic}

\usepackage{newfloat}
\usepackage{listings}
\DeclareCaptionStyle{ruled}{labelfont=normalfont,labelsep=colon,strut=off} 
\floatstyle{ruled}
\newfloat{listing}{tb}{lst}{}
\floatname{listing}{Listing}
\title{GreedyPrune: Retenting Critical Visual Token Set \\ for Large Vision Language Models}
\author{
    Ruiguang Pei\equalcontrib\textsuperscript{\rm 1}
    Weiqing Sun\equalcontrib\textsuperscript{\rm 2}\footnote{Work done during an internship at OPPO Research Institute},
    Zhihui Fu\textsuperscript{\rm 1},
    Jun Wang\textsuperscript{\rm 1}\footnote{Corresponding Author}
}
\affiliations{
    \textsuperscript{\rm 1}OPPO Reasearch Institute, Shenzhen, China\\
    \textsuperscript{\rm 2}Shanghai Jiao Tong University, Shanghai, China\\

    peiruiguang@gmail.com,
    sunweiqing@sjtu.edu.cn,
    luca@oppo.com,
    junwang.lu@gmail.com
    
}

\usepackage{bibentry}

\begin{document}

\maketitle

\begin{abstract}
Although Large Vision Language Models (LVLMs) have demonstrated remarkable performance in image understanding tasks, their computational efficiency remains a significant challenge, particularly on resource-constrained devices due to the high cost of processing large numbers of visual tokens.
Recently, training-free visual token pruning methods have gained popularity as a low-cost solution to this issue. However, existing approaches suffer from two key limitations: semantic saliency-based strategies primarily focus on high cross-attention visual tokens, often neglecting visual diversity, whereas visual diversity-based methods risk inadvertently discarding semantically important tokens, especially under high compression ratios.
In this paper, we introduce GreedyPrune, a training-free plug-and-play visual token pruning algorithm designed to jointly optimize semantic saliency and visual diversity. We formalize the token pruning process as a combinatorial optimization problem and demonstrate that greedy algorithms effectively balance computational efficiency with model accuracy.
Extensive experiments validate the effectiveness of our approach, showing that GreedyPrune achieves state-of-the-art accuracy across various multimodal tasks and models while significantly reducing end-to-end inference latency.
\end{abstract}

%

\begin{figure}[ht!]
    \centering
    \includegraphics[width=\columnwidth]{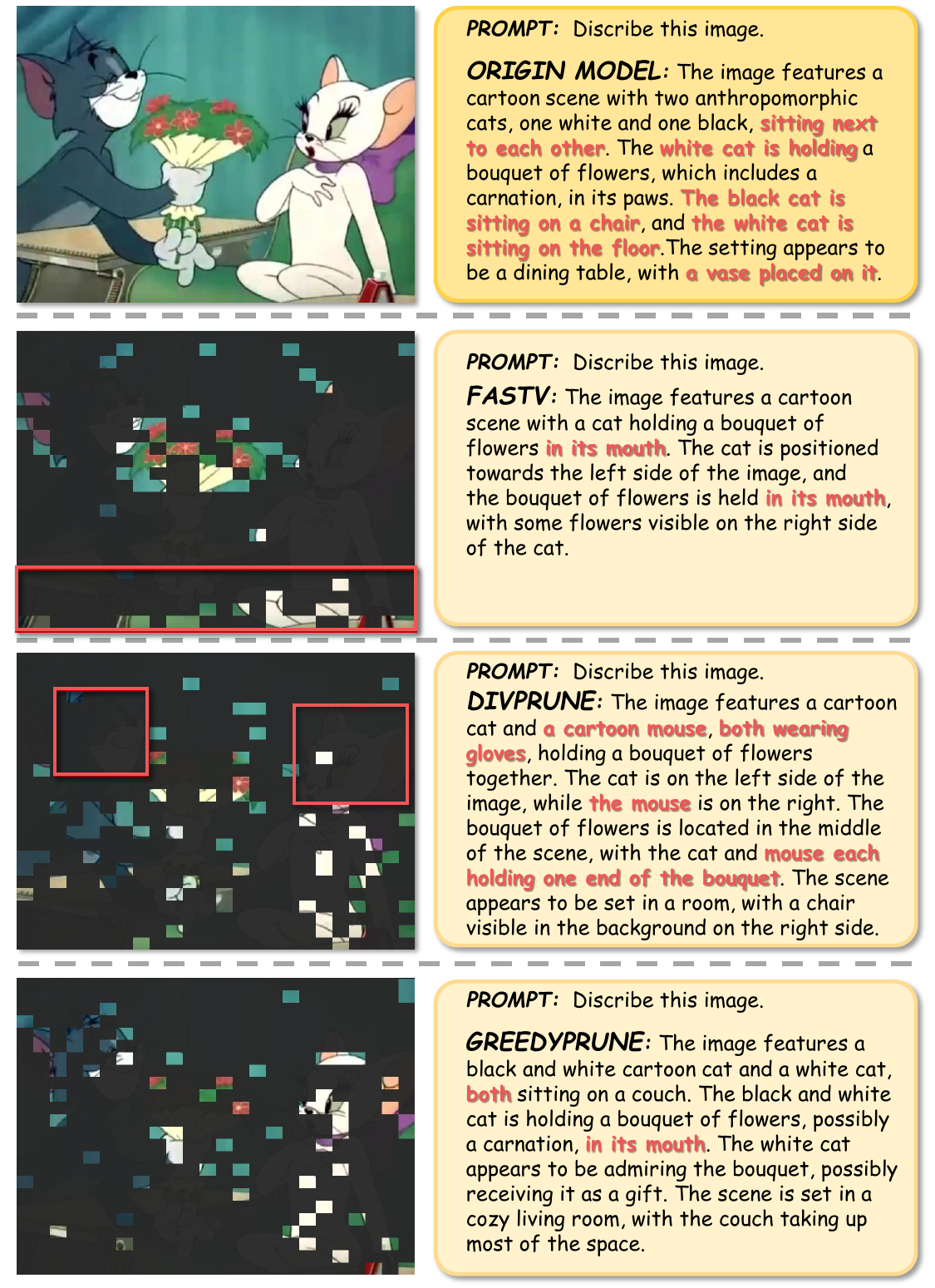}
    \caption{Comparison between baselines and GreedyPrune}
    \label{fig:fig1}
\end{figure}

\section{1 Introduction}


Large Vision Language Models (LVLMs) ~\cite{li_blip-2_2023, Liu_2024_CVPR_llava,chen2023sharegpt4v,bai_qwen25-vl_2025} have achieved remarkable success in image understanding tasks by deeply integrating visual information with natural language through the attention mechanism~\cite{vaswani_attention_2023}, However, the substantial computational cost and memory footprint associated with processing a large number of visual tokens present significant challenges for LVLM deployment on resource-constrained devices~\cite{jin2024efficientmultimodallargelanguage}.

To address this challenge, previous studies~\cite{bolya_token_2023,wu_ppt_2024, chen_image_2024,arif_hired_2024,alvar_divprune_2025,liu_multi-stage_2024,lin_boosting_2025,xing_pyramiddrop_2025,shang_llava-prumerge_2024,meng_plphp_2025} have investigated visual token pruning techniques that selectively remove or merge unimportant visual tokens at specific LVLM layers. This strategy reduces computational overhead by preventing unnecessary hidden state computations in subsequent layers, thereby minimizing end-to-end latency and reducing memory footprint.
Visual token pruning methods generally fall into two categories: training-based and training-free approaches. Training-based methods~\cite{yang2024visionzip,liang_dynamic_2025} preserve model accuracy even after token fusion or removal by leveraging fine-tuning or integrating new network components to effectively identify and eliminate redundant tokens. While these methods have shown promising results, they require additional training, which increases computational cost and may introduce potential performance degradation.


In training-free methods, approaches such as ~\cite{chen_image_2024, xing_pyramiddrop_2025, liu_multi-stage_2024} leverage cross-attention scores to evaluate the semantic saliency of visual tokens, enabling selective pruning. These methods, collectively referred to as \textbf{semantic saliency-based} approaches.
However, studies ~\cite{endo_feather_2024, zhang_cls_2024, wen_token_2025} have shown that cross attention guided visual token selection may introduce attention bias, as tokens located near the bottom of the image tend to be selected more frequently. This bias can lead to performance degradation, causing these approaches to underperform compared to random token selection in various tasks.
Another type of approaches ~\cite{wen_stop_2025, alvar_divprune_2025}, classified as \textbf{visual diversity-based method}, utilize clustering, similarity calculations, and other techniques to assess token diversity, effectively removing redundant tokens. These methods are founded on the observation that images inherently exhibit high redundancy and seek to retain a subset of tokens that maximize diversity. As a result, these approaches have achieved enhanced accuracy while maintaining the same visual token pruning ratio.

Although these approaches have achieved some success, we still observe the following issues: (1) The semantic saliency-based methods often retain redundant tokens within the selected subset, leading to a significant loss of image detail. Additionally, numerous studies have highlighted that these methods suffer from attention bias and lack compatibility with Flash Attention~\cite{dao2023flashattention2}, considerably reducing their practicality. (2) The visual diversity-based methods, when applied at high pruning ratios, tend to discard visual tokens with high semantic saliency, resulting in a rapid decline in accuracy under extreme pruning conditions.
Based on these insights, we reformulate the visual token pruning problem into a \textbf{critical visual token set} selection problem. This subset is designed to simultaneously preserve token diversity while maximizing the total semantic saliency value of the selected tokens.

In this paper, we introduce GreedyPrune, a novel tranning-free plug-and-play approach to perform visual token pruning. We formulate the selection of the critical visual token set as a combinatorial optimization problem and demonstrate that greedy algorithm strike an optimal balance between algorithmic efficiency and model accuracy. Across various scenarios, datasets, and LVLMs with equivalent pruning granularities, Extensive experimental validation demonstrates that GreedyPrune consistently outperforms existing advanced methods, achieving the highest accuracy while maintaining computational efficiency.
 For instance, on the LLaVA-1.5-13B model, even with an 88.9\% reduction of visual tokens, the average accuracy across 9 datasets remains an impressive 94.98\%, outperforming the second-best algorithm by 4.08\%.

Figure \ref{fig:fig1} presents a comparison between GreedyPrune and the baseline methods FastV and DivPrune. In each case, the left side illustrates the visualization of the retained visual token subset after pruning, while the right side shows the model’s response, where red-marked sections indicate hallucinations. The results reveal that while FastV identifies the main subject in an image, attention bias causes excessive token concentration at the bottom of the image, leading to the loss of critical visual information (e.g., "cat" and "coach"). Consequently, although FastV produces fewer hallucinations, the responses contain minimal useful information. On the other hand, DivPrune addresses this issue by distributing tokens more evenly with lower redundancy. However, it fails to account for the semantic significance of cross-modal visual tokens, causing the removal of key tokens representing image subjects, such as "cat," which leads to incorrect responses.


Main ontributions of this paper are as follows:
\begin{itemize}
    \item We reassessed the evaluation criteria for visual token semantic saliency and, through comprehensive quantitative and qualitative analysis, introduced a new, more effective metric.
    \item We propose that the selection of a critical visual token set is fundamental to visual token pruning. We reformulate this problem as a combinatorial optimization task and demonstrate that greedy algorithms can efficiently achieve near-optimal solutions.
    \item We introduce GreedyPrune, a training-free visual token pruning approach. Extensive experiments show that our method consistently achieves the highest accuracy across diverse scenarios, datasets, and LVLMs. In the LLaVA-1.5-7B model, GreedyPrune maintains 94.22\% accuracy even at an 88.9\% pruning ratio across 9 datasets, surpassing the second-best algorithm by 3.82\%.
\end{itemize}

\section{2 Related Work}

\subsection{2.1 Large Vision Language Model}
Leveraging visual encoders and large language models, Large Vision Language Models(LVLM)\cite{zhu_minigpt-4_2023,  Liu_2024_CVPR_llava,
wang_qwen2-vl_2024, bai_qwen25-vl_2025,zhao_accelerating_2024}integrate text and image modalities to enable cross-modal understanding and generation, significantly enhancing perception and reasoning capabilities in vision-related tasks.
Among open-source models, LLaVA-1.5-7B\cite{Liu_2024_CVPR_llava} employs a CLIPViT-MLP-LLM~\cite{dosovitskiy_image_2021} architecture for vision-language processing, achieving substantial performance improvements over Blip2\cite{li_blip-2_2023} and InstructBlip\cite{dai_instructblip_2023}. Models such as LLaVA-NeXT~\cite{liu2024llavanext}, Qwen2-VL\cite{wang_qwen2-vl_2024}, and Qwen2.5-VL\cite{bai_qwen25-vl_2025} support dynamic resolution input and multi-granularity image analysis, making them particularly suitable for high-resolution image processing. While these advancements lead to further performance improvements across various tasks, they also introduce higher computational costs.
\subsection{2.2 Efficient LVLM}
Several approaches have been proposed to enhance the efficiency of Large Vision Language Models (LVLMs). ~\cite{pmlr-v262-qiao24a} replaces the language transformer component of LVLMs with a more efficient module, achieving comparable performance with fewer parameters. ~\cite{xu2024llavadimattersmultimodallarge}leverages knowledge distillation to improve the capabilities of smaller models. Additionally, ~\cite{gagrani2024speculativedecodingmultimodallarge} extends speculative decoding techniques from large language models to LVLMs, significantly accelerating the decoding process.
Among these methods, visual token pruning has garnered extensive research attention. In training-based approaches, ~\cite{yang2024visionzip} perform token pruning or fusion is followed by fine-tuning the network to adapt to the new representations, ensuring accuracy retention. \cite{chen_efficient_2024} train visual context compressors to compact redundant tokens, \cite{liang_dynamic_2025} fine-tune the base model to induce token sparsity during prefill, subsequently training a predictor to apply sparsity to decoded tokens. However, such approaches above often necessitate additional training or modifications to existing models, incurring considerable computational costs and potentially leading to accuracy degradation. The work on training-free vision token pruning will be introduced in detail in the next section.
\subsection{2.3 Training-free Visual Token Prune}

Training-free visual token pruning primarily operates within specific layers of the LLM component in LVLM, where visual tokens are pruned according to predefined metrics. The pruned tokens are excluded from subsequent computations, thereby improving the efficiency of LVLM.
FastV\cite{chen_image_2024} employs a shallow-layer pruning strategy, utilizing the cross-attention of the last text token in the input sequence to select the top-K visual tokens. VTM\cite{lin_boosting_2025} performs mid-layer pruning of all visual tokens, sometimes even enhancing accuracy in certain tasks. PyramidDrop\cite{xing_pyramiddrop_2025} adopts a multi-stage pruning approach, progressively reducing the number of tokens as layers deepen, while MustDrop\cite{liu_multi-stage_2024} further refines the pruning strategy by considering pruning at three distinct stages: visual encoding, prefill, and decode.
FasterVLM\cite{zhang_cls_2024} identifies attention bias in cross-attention selection criteria (primarily caused by ROPE properties) and instead utilizes [CLS] token attention as the selection metric. Since the aforementioned methods primarily rely on attention mechanisms to gain the semantic saliency for visual token pruning, they can be collectively categorized as \textbf{semantic saliency-based methods}.
Recent studies\cite{wen_token_2025} suggest that attention may not be the optimal selection criterion. Instead, reducing redundancy while ensuring token diversity is more critical. Based on this, Dart\cite{wen_stop_2025} removes token redundancy using cosine similarity, achieving better accuracy improvements compared to prior methods at the same pruning ratio. Divprune\cite{alvar_divprune_2025} further refines pruning by applying the Min-Max-Diversity approach to eliminate token redundancy. These methods can be categorized as \textbf{visual diversity-based methods}.
\section{3 Methodology}

\begin{figure}[ht]
    \centering
    \begin{subfigure}{0.3\columnwidth}
        \centering
        \includegraphics[width=\linewidth]{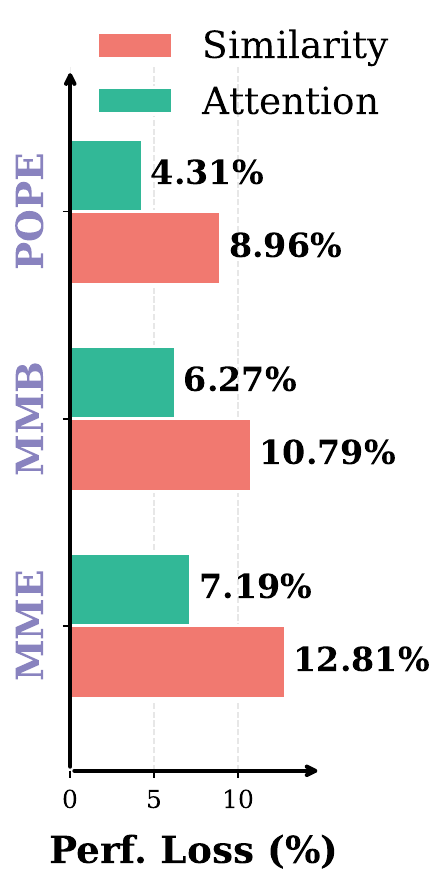}
        \caption{\textbf{Metric comparison}: performance loss caused by the removal of top 20\% of visual tokens selected based on the showed metrics.}
        \label{fig:sm_sub1}
    \end{subfigure}
    \hfill
    \begin{subfigure}{0.69\columnwidth}
        \centering
        \includegraphics[width=\linewidth]{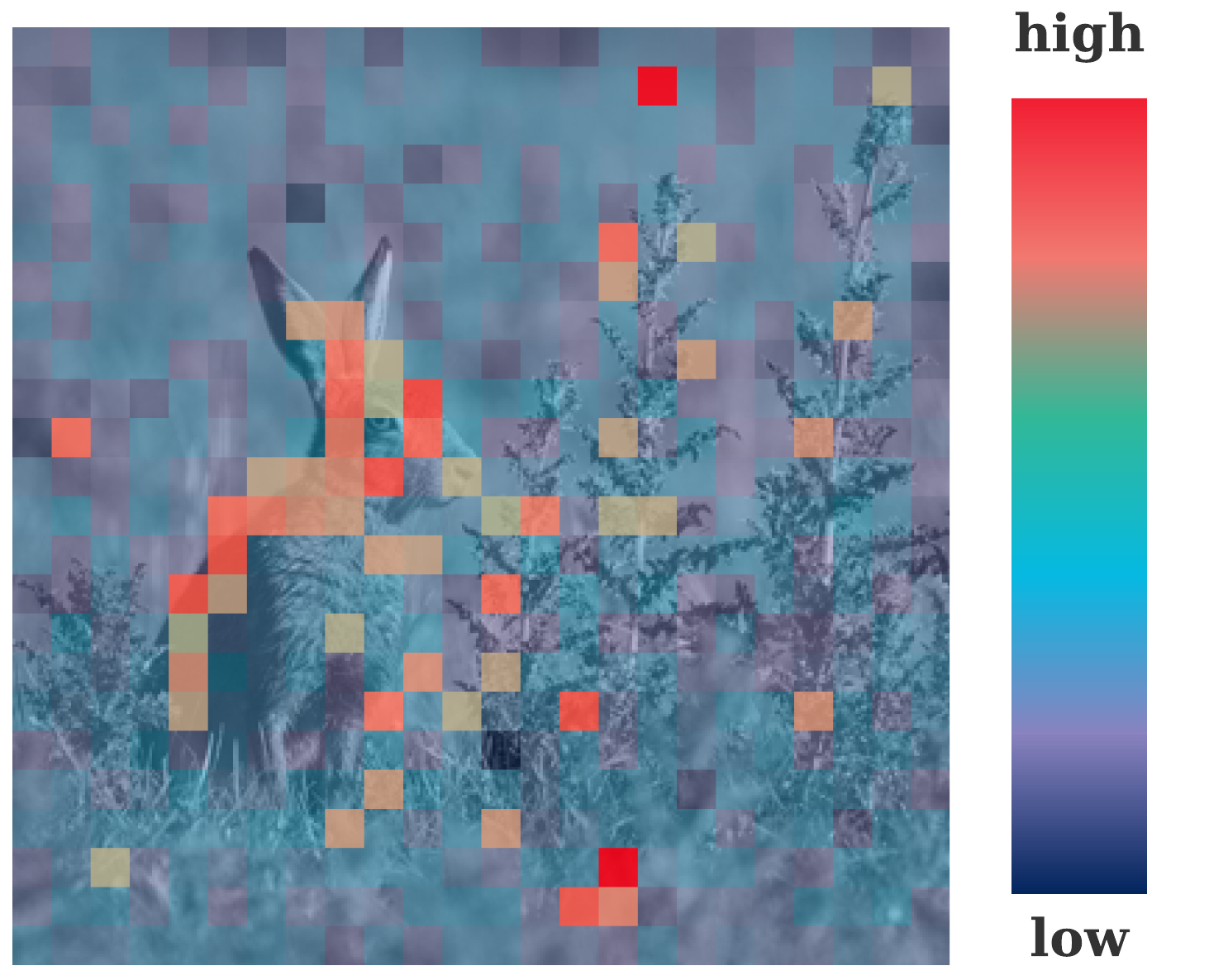}
        \caption{\textbf{Visualization of the cosine similarity heatmap between the last text token and visual tokens} reveals that the metric effectively captures key semantic regions within the image, such as rabbits, plants, and grassy fields. Notably, it maintains accurate attention distribution without exhibiting any bias.}
        \label{fig:sm_sub2}
    \end{subfigure}
    \caption{Rethinking semantic saliancy of visual token}
    \label{fig:sm_all}
\end{figure}
Given the inherent limitations of semantic saliency and diversity-based approaches, this paper formulates the problem of training-free visual token pruning as a combinatorial optimization task focused on selecting an optimal subset of visual tokens. Specifically, under the constraint with ensuring high diversity (minimizing redundancy) among the fixed subset size of selected tokens. This chapter begins with an analysis of the key metrics for assessing the semantic saliency of visual tokens, followed by a detailed formulation and theoretical examination of the subset selection problem, and concludes with a comprehensive presentation of the proposed methodology.

\subsection{3.1 Rethinking Semantic Saliency Of Visual Token}
%

Contemporary methods such as ~\cite{chen_image_2024} and ~\cite{xing_pyramiddrop_2025} rely on the attention score of the last text token on visual tokens to evaluate semantic saliency and select token subsets. However, studies like ~\cite{zhang_cls_2024} and ~\cite{endo_feather_2024} highlight an inherent bias in this approach that visual tokens closer to text tokens tend to receive greater attention, a phenomenon attributed to the properties of ROPE~\cite{su2023roformerenhancedtransformerrotary}.

We revisit this issue and propose a more effective way to determine visual token semantic saliency. Prior research~\cite{zhang2024cross} has shown that LVLM exhibits information flow within its LLM component during inference. In the shallow layers of LLM, global image token information propagates into the text token. Drawing inspiration from this observation and FastV~\cite{chen_image_2024}, we introduce cosine similarity between the last text token and visual tokens as an alternative metric for assessing semantic saliency. This approach offers several advantages:(1)normalization by vector magnitude, which may mitigate the influence of positional encoding on semantic similarity evaluation.(2)independence from attention scores, allowing compatibility with Flash Attention mechanisms.(3)scalable to any LVLMs.

As illustrated in Figure ~\ref{fig:sm_sub1}, we conduct a metric comparison study using the LLaVA 1.5-7B model across multiple datasets, including MME, POPE, and MMB, to compare cosine similarity against traditional attention score-based evaluations. Specifically, we compute the semantic saliency of all visual tokens using two distinct methods and remove the top 20\% most "valuable" tokens. The experimental results, shown in the figure, indicate that tokens selected based on cosine similarity have a more decisive impact on task accuracy compared to those identified via attention score-based approaches. The removal of these tokens leads to a more pronounced performance decline.

Additionally, we visualize the cosine similarity metric. As depicted in Figure ~\ref{fig:sm_sub2}, this metric effectively identifies semantically significant regions within images, ensuring robust attention distribution without introducing attention shift artifacts.

\subsection{3.2 Problem Formulation and Theoretical Analysis}

%
Let $\mathcal{V} = \{v_1, v_2, \dots, v_n\}$ denote the original set of visual tokens, where the total number of tokens is $N$. Each token is associated with a semantic saliency weight, represented as $\mathcal{W} = \{w_1, w_2, \dots, w_n\}$. This study aims to select a subset of tokens $\mathcal{S}$, with a subset size of $M$, where $M \ll N$, to enhance the inference efficiency of LVLM. Furthermore, we seek to maximize the semantic saliency of the selected token subset while ensuring a high degree of diversity (low redundancy). The optimization problem can be formulated as follows:

\begin{equation}
\begin{aligned}
\max_{\mathcal{S} \subseteq \mathcal{V}} & \sum_{v_i \in \mathcal{S}} w_i \\
\text{s.t.} \quad \cos(v_i, v_j) \leq \tau&, \quad \forall v_i, v_j \in \mathcal{V}, \, i \neq j
\end{aligned}
\label{eq:optimization}
\end{equation}

In this optimization problem, $\cos(x_i, x_j)$ represents the cosine similarity between elements $v_i$ and $v_j$, which must satisfy the threshold $\tau$ to ensure sufficient diversity within the subset.

To facilitate the formulation, we introduce the binary variable $z_i \in \{0,1\}$, which denotes whether the element $v_i$ is selected. Based on this definition, the problem is reformulated as follows:
\begin{equation}
\begin{aligned}
\max_{z \in \{0,1\}^n} & \sum_{i} w_iz_i \\
\text{s.t.} \quad (\cos(v_i, v_j) -& \tau)z_iz_j\leq0, \quad \forall i<j
\end{aligned}
\label{eq:optimization2}
\end{equation}

For each set of constraints, we incorporate the non-negative Lagrange multipliers $\lambda_{i,j} \geq 0$, thereby constructing the corresponding Lagrangian function to systematically represent the constraints and optimize the objective function.
\begin{equation}
L(z, \lambda) = \sum_{i} w_i z_i - \sum_{i<j} \lambda_{i,j} (\cos(v_i, v_j) - \tau) z_i z_j
\label{eq:lagrangian}
\end{equation}

The overall optimization problem can be reformulated as follows.
\begin{equation}
\max_{z \in \{0,1\}^n} L(z, \lambda)
\label{eq:relaxed_subproblem}
\end{equation}

In equation~\eqref{eq:lagrangian}, the first term represents the linear component, while the second term corresponds to the quadratic component. The values of $z_i$ and $z_j$ are restricted to $\{0, 1\}$, and the cosine similarity matrix exhibits symmetry. As a result, this problem falls within the domain of Unconstrained Binary Quadratic Programming (UBQP)~\cite{kochenberger2014unconstrained}, which has been proven to be NP-hard. Although exact solutions exist in theory, their computational complexity grows exponentially. For instance, the branch-and-bound method can be employed to derive an exact solution; however, under unfavorable conditions, its complexity may reach $2^n$.

In practical applications, heuristic methods provide approximate solutions within a manageable time complexity, encompassing approaches such as greedy algorithm.
During each iteration of the greedy algorithm, a locally optimal solution is derived by adhering to the following constraints:

\begin{equation}
\max_{i} w_i - \lambda_{i,j} (\cos(v_i,v_j) -\tau), \quad\forall i<j 
\label{eq:greedy_constraint}
\end{equation}

At each step, the visual token with the highest current value of $w_i$is identified, while highly correlated elements are systematically eliminated.


\begin{figure*}[t!]
    \centering
    \includegraphics[width=\textwidth]{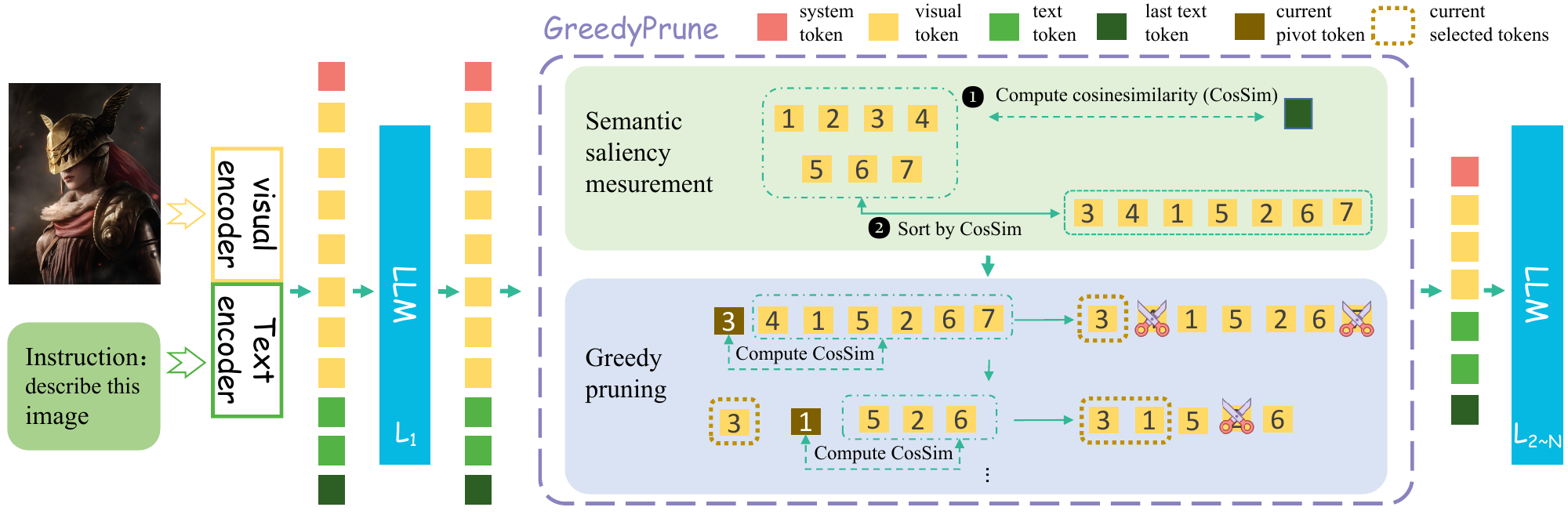}
    \caption{An overview of the architecture of LVLM with GreedyPrune}
    \label{fig:main}
\end{figure*}

\subsection{3.3 GreedyPrune}

Building on the above analysis, we redefine the visual token pruning problem as ~\eqref{eq:optimization}, whose exact solution has been proven to be NP-hard. Since this study focuses on efficiency in LVLM inference, we employ a heuristic algorithm with relatively low computational complexity to obtain an approximate solution.
To this end, we introduce a greedy token pruning method (fig.~\ref{fig:main}), which selects an appropriate token subset while ensuring that each iteration step of the algorithm satisfies the constraint ~\eqref{eq:greedy_constraint}. Our approach is training-free, plug-and-play, and model-agnostic, making it applicable to any LVLMs.


As illustrated in fig.~\ref{fig:main}, the proposed method is integrated between the first and second layer of the LLM component within the LVLM architecture. It consists of two primary steps: first, semantic saliency measurement of visual tokens; second, iterative greedy pruning.
In the first step, we compute the cosine similarity ($CosSim$) between each visual token and the last text token to derive its semantic value. The visual tokens are then sorted in descending order based on $CosSim$. Naturally, visual tokens with higher semantic values exhibit greater semantic saliency.
During the iterative greedy pruning process, let $V_{sorted}$ denote the current set of tokens to be pruned and $V_{critic}$ represent the selected critical visual token set. In each greedy iteration, the first element of $V_{sorted}$ is extracted as the pivot token, denoted as $v_{pivot}$, and the sets are updated as follows: $V_{sorted} = V_{sorted} \setminus \{ v_{pivot}\} $, $V_{critic} = V_{critic} \cup \{ v_{pivot}\} $.
Next, the pairwise similarity $CosSim$ between $v_{pivot}$ and each remaining element in $V_{sorted}$ is computed. A subset, $V_{redundancy}$, is formed by selecting elements with similarity exceeding the threshold $\tau$, and these redundant tokens are subsequently removed from $V_{sorted}$, updating the set as $V_{sorted} = V_{sorted} \setminus V_{redundancy}$.
This process continues until $V_{critic}$ reaches the predefined subset size $M$, or until $V_{sorted}$ becomes empty.
At each iteration, the algorithm consistently selects the visual token with the highest semantic saliency while eliminating highly redundant tokens, thereby ensuring that the greedy pruning method satisfies the constraint~\eqref{eq:greedy_constraint}.

\section{4 Experiment}
In this section, we present a comprehensive performance evaluation of the proposed method in comparison with previous works, considering diverse settings, datasets, tasks, and LVLM models. Furthermore, we conduct ablation studies to systematically assess the impact of various factors on the effectiveness of our approach.
\subsection{4.1 Experimental Setings}
All experiments in this study were conducted using 8 × H20 GPUs with 120GB of VRAM. For benchmarking across various tasks and datasets, including both baseline models and the proposed approach, we employed the lmms-evals~\cite{zhang2024lmmsevalrealitycheckevaluation, lmms_eval2024} package. Throughout all experiments, the batch size was consistently set to 1.
\begin{table*}[ht!]
\centering
\caption{Comparative experiments on image understanding tasks. In all our experiments, $\tau$ represents the similarity threshold between image tokens, and K denotes the starting layer for token pruning. The metrics listed below each dataset correspond to the evaluation criteria for the respective task.The bold numbers indicate that the method achieves the best performance, and the underlined numbers indicate that the method achieves the sub-optimal performance.}
\label{tab:result_with_same_token_budget}
\resizebox{\textwidth}{!}{
    \begin{tabular}{c|c|cc|ccccccccc|c}
        \toprule[2pt]
        \multirow{2}{*}{Model} & \multirow{2}{*}{Method} & \multirow{2}{*}{K} & \multirow{2}{*}{\makecell{Retained \\tokens}} & MME & POPE & MMB & Ocrbench & TextVQA & OK-VQA & Nocaps & Flickr30K & GQA & \multirow{2}{*}{Avg.} \\
        ~ & ~ & ~ & ~ & P-score & F1 & Acc & Acc & EM & EM & CIDEr & CIDEr & EM & ~\\ 
        \midrule
        
        \multirow{13}{*}{\rotatebox{90}{LLaVA-1.5-7B} }&\cellcolor{gray!30} Original & \cellcolor{gray!30}- & \cellcolor{gray!30}576 & \cellcolor{gray!30}1509 & \cellcolor{gray!30}85.8 &\cellcolor{gray!30} 64 & \cellcolor{gray!30}314 &\cellcolor{gray!30} 46 &\cellcolor{gray!30} 53.4 &\cellcolor{gray!30} 105.5 &\cellcolor{gray!30} 74.8 &\cellcolor{gray!30} 61.9 &\cellcolor{gray!30} 100\% \\ 
        \cmidrule{2-14}
        \rule{0pt}{10.5pt}~ & FastV & 3 & 192    & 1475             & 79.7              & \textbf{63.6}& 298            & 44                             & \underline{51.7}           & \textbf{104.2}& \underline{73.3}          & 58.6 & 96.54\% \\ 
        \rule{0pt}{10.5pt}~ & DART & 2 & 192     & \textbf{1493}    & 82.7            & 63.2            & \textbf{305}   & \underline{44.3}                 & 51.4           & \underline{103.7}   & \textbf{74.3} & \underline{59.9} & 97.57\% \\ 
        \rule{0pt}{10.5pt}~ & DivPrune & 0 & 192 & 1426             & \textbf{87.1}   & 62.4            & \underline{303}& 43.1                             & 51.6  & 100.4            & 71.5          & \underline{59.9} & 96.43\% \\ 
        \rule{0pt}{10.5pt}~ & Ours($\tau$=0.78) & 1 & 192     & \underline{1488} & \underline{85.5}& \underline{63.3}   & 301   & \textbf{44.4}                    &\textbf{51.9}& 102.1            & 73& \textbf{61.4} & \textbf{97.81\%} \\
        \cmidrule{2-14}
        \rule{0pt}{10.5pt}~ & FastV & 3 & 128 & 1450 & 76.1 & \underline{62.4} & 286 & 42 & 50.3 & \underline{101.2} & 70.8 & 56.8 &  93.47\% \\ 
        \rule{0pt}{10.5pt}~ & DART & 2 & 128 & \underline{1468} & 79.7 & \textbf{62.5} & \underline{287} & \underline{42.5} & \underline{50.8} & \textbf{101.8} & \textbf{73.2} & 58.6 & 95.09\% \\
        \rule{0pt}{10.5pt}~ & DivPrune & 0 & 128 & 1401 & \textbf{86.7} & 62.2 & 283 & 41.9 & 49.9 & 98 & 69.2 & \underline{59.4} & 94.12\% \\ 
        \rule{0pt}{10.5pt}~ & Ours($\tau$=0.86)& 1 & 128 & \textbf{1483} & \underline{85.6} & \underline{62.4} & \textbf{291} & \textbf{43.6} & \textbf{51.2} & \textbf{101.8} & \underline{72.2} & \textbf{61.2} & \textbf{96.75\%} \\ 
        \cmidrule{2-14}
        \rule{0pt}{10.5pt}~ & FastV & 3 & 64 & 1363 & 67.9 & 59.5& 239 & 36.9 & 47 & 90.6& 62.4 & 53.4 & 84.71\% \\ 
        \rule{0pt}{10.5pt}~ & DART & 2 & 64 & 1390 & 72.7 & \underline{59.9} & 260 & 37.8 & 47.7 & \underline{96.9} & \underline{68.6} & 55.6 & 88.68\% \\ 
        \rule{0pt}{10.5pt}~ & DivPrune & 0 & 64 & \underline{1368} & \textbf{85.6}  & 59.4  & \underline{275} & \underline{39.3}  & \underline{47.9}  & 93.6  & 64.4  & \underline{57.6}  & 90.42\% \\ 
        \rule{0pt}{10.5pt}~ & Ours($\tau$=0.94) & 1 & 64 & \textbf{1442} & \underline{84.4}  & \textbf{61.3}  & \textbf{286}  & \textbf{42}  & \textbf{49.2}  & \textbf{98.4}  & \textbf{69.5}  & \textbf{60.4}  & \textbf{94.22\%} \\ 
        \midrule
        \rule{0pt}{10.5pt}\multirow{5}{*}{\rotatebox{90}{\small LLaVA-1.5-13B}} & \cellcolor{gray!30}Original & \cellcolor{gray!30}- & \cellcolor{gray!30}576 & \cellcolor{gray!30}1530 &\cellcolor{gray!30} 85.9 &\cellcolor{gray!30} 68.6 &\cellcolor{gray!30} 338 &\cellcolor{gray!30} 48.7 &\cellcolor{gray!30} 58.2 &\cellcolor{gray!30} 109.3 &\cellcolor{gray!30} 79.4 &\cellcolor{gray!30} 63.2 & \cellcolor{gray!30}100\%\\ 
        \cmidrule{2-14}
        \rule{0pt}{10.5pt}~ & FastV & 3 & 64 & 1360 & 70.8 & 63.8 & 235 & 36.6 & 51.4 & 96.3 & 67.7 & 56.1 & 84.38\% \\ 
        \rule{0pt}{10.5pt}~ & DART & 2 & 64 & 1418 & 73.9 & \underline{65.1} & 266 & 35.1 & 51.9 & \underline{100.7} & \underline{73.2} & 56.3 & 87.23\% \\ 
        \rule{0pt}{10.5pt}~ & DivPrune&0 & 64 & \underline{1473} & \underline{84.6} & 64.7 & \underline{297} & \underline{39.7} & \underline{54.2} & 97.6 & 68.2 & \underline{57.7} & 90.90\% \\ 
        \rule{0pt}{10.5pt}~ & Ours($\tau$=0.78)&1 & 64 & \textbf{1524} & \textbf{85} & \textbf{65.5} & \textbf{323} & \textbf{45.3} & \textbf{55.5} & \textbf{102.4} & \textbf{74.5} & \textbf{61.6} & \textbf{94.98\%} \\
        \midrule
        \multirow{5}{*}{\rotatebox{90}{\small LLaVA-1.6-7B}} &\cellcolor{gray!30} Original & \cellcolor{gray!30}- &\cellcolor{gray!30} -& \cellcolor{gray!30}1519 & \cellcolor{gray!30}86.4 &\cellcolor{gray!30} 67.1 &\cellcolor{gray!30} 521 &\cellcolor{gray!30} 64.8 & \cellcolor{gray!30}44.2 & \cellcolor{gray!30}88.3 &\cellcolor{gray!30} 68.4 &\cellcolor{gray!30} 64.2 & \cellcolor{gray!30}100\%\\ 
        \cmidrule{2-14}
        \rule{0pt}{10.5pt}~ & FastV & 3 & 320 & 1410 & 79.6 & \underline{64.6} & 375 & 52.4 & 40.1 & 77.2 & 59.7& 59.1 & 87.95\% \\ 
        \rule{0pt}{10.5pt}~ & DART & 2 & 320 & 1424 & 83.3 & 64.3 & \textbf{408} & \textbf{58.2} & 41.5 &\textbf{81.6} & 62.8 & \underline{61.3} & 91.97\% \\ 
        \rule{0pt}{10.5pt}~ & DivPrune & 0 & 320 & \textbf{1441} & \underline{84.1} & \textbf{64.8} &347 & 51.1 & \underline{43.2} & 79.9 & \textbf{64.2} & 60.4 & 90.04\% \\ 
        \rule{0pt}{10.5pt}~ & Ours($\tau$=0.90)&1 & 320 & \textbf{1441} & \textbf{85.7} & 63.1 & \underline{396} & \underline{56.5} & \textbf{44.4} & \underline{80.8} & \underline{63.1} & \textbf{62} & \textbf{92.45\%} \\
        \bottomrule[2pt]
    \end{tabular}
}
\end{table*}
\subsubsection{Baselines}
We evaluate three baseline methods: FastV~\cite{chen_image_2024}, DART~\cite{wen_stop_2025}, and DivPrune~\cite{alvar_divprune_2025}. These methods are our main competitors as they are plug-and-play solutions that do not require additional fine-tuning or calibration. FastV are semanti saliency-based, while dart and Divprune follow a visual diversity-based approaches.
To maintain consistency, we use the default parameters from their respective papers and open-source implementations. FastV prunes tokens at layer $K=3$, dart at layer 2, and Divprune at layer 0.
\subsubsection{Models}

To assess the generalizability of greedy prune in LVLMs, we evaluate its performance across multiple models, including LLaVA-1.5-7B, LLaVA-1.5-13B, LLaVA-1.6-7B(LLaVa-NEXT-7B), in image-based scenarios.
For each model and task, we report only the relevant subset of baselines alongside our results. 

In LLaVA-series models, the visual encoder is based on CLIP-vit. Specifically, LLaVA-1.5 encodes images into 576 tokens using a fixed resolution, while LLaVA-NEXT is optimized for high-resolution image processing, segmenting images into multiple parts and encoding them into 2880 tokens.

\subsubsection{Datasets, Tasks and Metrics}

We selected 9 image task datasets, covering most image understanding tasks. In general visual question answering, we chose POPE~\cite{li-etal-2023-evaluating-pope}, MME~\cite{fu2023mme}, MMB~\cite{liu2024mmbenchmultimodalmodelallaround}, OK-VQA~\cite{okvqa} and GQA~\cite{hudson2018gqa}. For text-related visual question answering, we included orcbench~\cite{Liu_2024_ocrbench} and TextVQA~\cite{text_vqa}. In image captioning, we utilized Nocaps~\cite{agrawal2019nocaps} and Flickr30K~\cite{plummer2016flickr30kentitiescollectingregiontophrase}.

In these datasets, LVLM generates close-ended responses based on predefined prompts. To ensure consistency with previous work, we use the CIDEr score as the evaluation metric for image captioning. For other close-ended question-answering tasks, metrics such as Exact Match (EM), Accuracy (Acc), F1, and Perception Score (P-score) are used. Detailed evaluation results are provided in later sections.


\subsection{4.2 Image Understanding Experiment Results}


\subsubsection{Aligned token pruning ratio}

The experimental results are shown in Table \ref{tab:result_with_same_token_budget}. To ensure a fair comparison, we first aligned the token pruning ratios of all methods across all datasets, allowing us to evaluate their accuracy under the \textbf{same memory constraints}.
We conducted pruning experiments on the LLaVA-1.5-7B model at various pruning ratios. At a 66.7\% pruning ratio (retaining 192 visual tokens), our approach achieved an overall average accuracy of 97.81\%, outperforming the second-best algorithm by 0.36\%. As the pruning ratio increased, the performance of competing methods dropped sharply, whereas our approach exhibited gradual performance degradation, demonstrating superior robustness.
At a 77.8\% pruning ratio (retaining 128 visual tokens), our method ranked second on the POPE dataset and two caption datasets, while achieving first place on all other datasets, with an average accuracy of 96.75\%, exceeding the second-best method by 2.66\%.
At an 88.9\% pruning ratio (retaining 64 visual tokens), FastV and DART experienced a rapid decline in accuracy, whereas our method maintained first place on all datasets except POPE, where it ranked second. Despite the extreme pruning ratio, our method retained an average accuracy of 94.22\%, surpassing the second-best approach by 3.82\%. These results highlight the strong resilience of our approach which performance degradation remains slow even under high pruning ratios, making it practically viable for extreme compression scenarios.


We further conducted 88.9\% pruning ratio experiments on LLaVA-1.5-13B and LLaVA-1.6-7B (LLaVA-NEXT-7B) to validate our approach across different model scales. On LLaVA-1.5-13B, our method achieved first place across all tasks, with an overall average accuracy of 94.98\%, outperforming the second-best approach by 4.08\%.
On LLaVA-NEXT-7B, GreedyPrune continued to demonstrate exceptional performance, highlighting its scalability and generalization ability across different model sizes. 
\begin{figure*}[htbp] 
  \centering 
  \begin{subfigure}{0.32\textwidth} 
    \includegraphics[width=\linewidth]{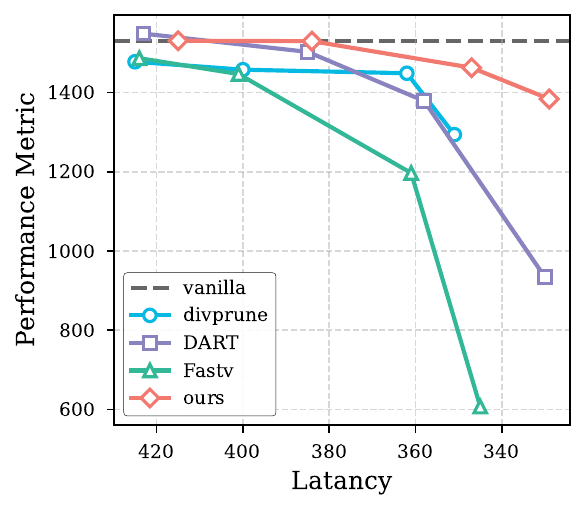}
    \caption{MME}
    \label{fig:sub1}
  \end{subfigure}
  \hfill 
  \begin{subfigure}{0.32\textwidth}
    \includegraphics[width=\linewidth]{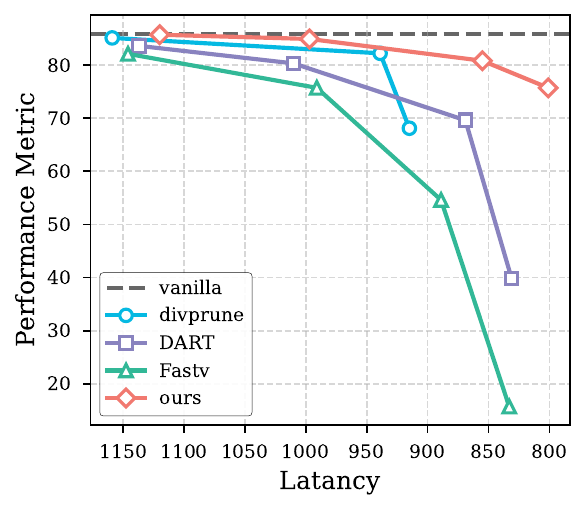}
    \caption{POPE}
    \label{fig:sub2}
  \end{subfigure}
  \hfill
  \begin{subfigure}{0.32\textwidth}
    \includegraphics[width=\linewidth]{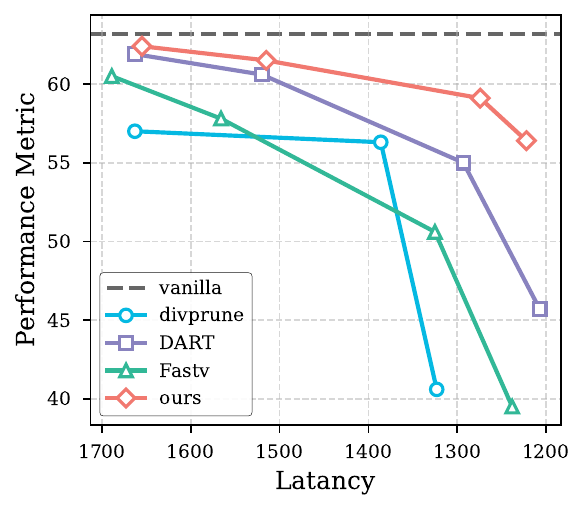}
    \caption{GQA}
    \label{fig:sub3}
  \end{subfigure}
  \caption{Performance-Latency trade-off comparisons across different datasets on LLaVA-1.5-13B.}
  \label{fig:three_figures}
\end{figure*}
\begin{table*}[ht!]
\centering
\caption{Ablation study of semantic saliency and greedy redundancy removal}
\label{tab:aba1}
\resizebox{0.85\textwidth}{!}{
    \begin{tabular}{c|cc|ccccc}
        \toprule[1.5pt]
        LLava-1.5-7B & Semantic saliency & Redundancy removal  & MME & POPE & TextVQA & GQA &  Nocaps \\
        \midrule
        \rowcolor{gray!30}
        origin &- &- & 1509 & 85.8	& 46	& 61.9	& 105.5 \\
        (a) &\ding{56}& \ding{56} & 1275	& 71.5	& 21.2 & 56.1 &	93.2 \\
        (b) &\ding{52} & \ding{56}&1384	&73.3	&38.2	&56.4	&97.1 \\
        (c) &\ding{56} &\ding{52}&1209&	70.8	&20.6	&54.8	&88 \\
        GreedyPrune &\ding{52} & \ding{52}&\textbf{1427}&\textbf{84.5} &\textbf{41.9}	&\textbf{60.2}	&\textbf{98.4} \\
        \bottomrule[1.5pt]

    \end{tabular}
    }\label{table1}
\end{table*}
\subsubsection{Performance-Latency trade-off}
We conducted a comprehensive evaluation of the trade-offs between model accuracy and end-to-end inference latency on the LLaVA-1.5-13B model, using the POPE, MME, and GQA datasets. To ensure consistency and fairness, all experimental cases were executed on a single GPU, with model accuracy and runtime latency measured via the lmms-eval evaluation framework.
As illustrated in Figure \ref{fig:three_figures}, our method consistently achieved superior performance across diverse datasets and latency requirements. A key observation is that when the pruning ratio increases in low-latency settings, competing algorithms experience a sharp decline in model accuracy, often struggling to maintain acceptable performance levels. In contrast, GreedyPrune exhibits strong resilience, retaining high accuracy while undergoing slower degradation, ultimately outperforming other methods across all tested scenarios.
Our approach strikes an optimal balance between end-to-end inference latency and model accuracy, systematically surpassing existing techniques. This advancement is driven by two fundamental innovations: an optimized semantic saliency selection metric for visual tokens and a greedy token redundancy pruning strategy. Together, these refinements minimize unnecessary visual feature processing, leading to a significant boost in benchmark performance for vision-language understanding tasks.

\subsection{4.3 Abalation Study}
\begin{figure}[ht!]
    \centering
    \includegraphics[width=\columnwidth]{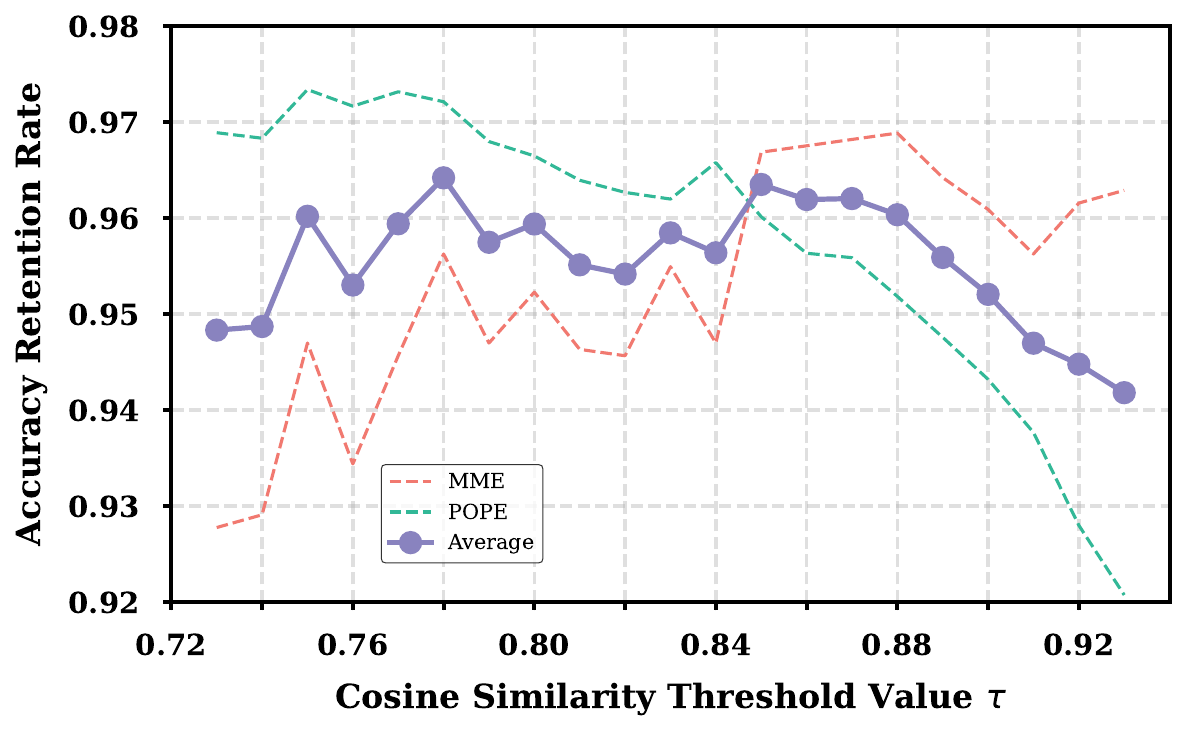}
    \caption{ablation study of similarity value $\tau$}
    \label{fig:aba_tau}
\end{figure}

This section first conducts ablation experiments on the two primary optimization strategies proposed in this paper which are semantic saliency metric and greedy redundancy removal. Subsequently, we further analyze the impact of the redundancy threshold $\tau$, with results presented in Table \ref{tab:aba1} and Figure ~\ref{fig:aba_tau}. All ablation studies were conducted using the LLaVA-1.5-7B model, with 88.9\% of visual tokens pruned.
\subsubsection{Ablation study of Semantic saliency}
We first implemented two simple algorithms, as shown in (a) and (b) in Table \ref{tab:aba1}. Specifically, (a) employs a random selection approach, whereas (b) leverages our proposed semantic saliency metric to rank visual tokens and select the most critical token set. A comparative analysis of (a) and (b) reveals that method (b) significantly enhances overall performance compared to random selection (a), thereby validating its effectiveness. Moreover, as observed in Table ~\ref{tab:result_with_same_token_budget}, method (b) demonstrates substantially superior performance over the similar approach FastV, further confirming that our proposed metric surpasses the cross-attention score in identifying visual tokens with high semantic saliency. Additionally, since this metric operates independently of attention scores, it is fully compatible with flash-attention.

\subsubsection{Ablation study of greedy redundancy removal}
We further implemented an algorithm that exclusively incorporates greedy redundancy removal, denoted as (c) in Table ~\ref{tab:aba1}. This algorithm first applies a random shuffle to the visual tokens before executing the greedy redundancy removal process. At each step, it selects the first element from the candidate subset, adds it to the result subset, and computes its similarity with the remaining elements in the candidate subset, discarding those with high cosine similarity. Compared to GreedyPrune, the only modification in algorithm (c) is the replacement of semantic saliency-based sorting with random sorting in the initial step. Experimental results indicate that this approach performs comparably to random selection or even slightly worse. This suggests that under random ordering, greedy redundancy removal is ineffective at eliminating redundancy and cannot guarantee a locally optimal solution at each step. However, when semantic saliency sorting is combined with greedy redundancy removal, which forming our proposed method GreedyPrune, the greedy redundancy removal mechanism significantly enhances the results obtained through semantic saliency-based selection of critical visual tokens. This approach consistently achieves the best performance across all benchmarks, particularly on the POPE dataset, thereby demonstrating the effectiveness of the optimization strategies employed in GreedyPrune.

\subsubsection{ablation study of similarity value $\tau$}
The cosine similarity value $\tau$ serves as a crucial balancing metric. Specifically, when $\tau$ is lower, redundancy evaluation becomes more stringent, leading the critical visual token set to favor visual tokens with lower redundancy, as the pairwise similarity within the subset must not exceed $\tau$. Conversely, when $\tau$ is higher, redundancy evaluation is more permissive, allowing the critical visual token set to prioritize visual tokens with higher semantic saliency. Using the LLaVA-1.5-7B model with a pruning intensity of 88.9\%, we examine the impact of varying $\tau$ on the MME and POPE datasets. As shown in Figure 3, the red and green dashed lines represent the accuracy ratios of the MME and POPE datasets, respectively. As $\tau$ increases from 0.73 to 0.93, POPE initially improves, peaking at 0.78 before sharply declining, while MME continues to rise. These findings suggest that different datasets exhibit varying preferences regarding the redundancy level of the visual token set, which some favor lower redundancy, whereas others prioritize higher semantic saliency. The purple curve in Figure 3, representing the average accuracy ratio across datasets, first increases and then declines, suggesting the existence of an optimal $\tau$ that balances performance across visual understanding tasks. In practical applications, this parameter can be fine-tuned to achieve optimal performance tailored to specific scenarios.
\section{5 Conclusion}
In this paper, we address the core challenge of training-free visual token pruning is to identify a critical visual token set that simultaneously preserves visual diversity and semantic saliency. We formulate this problem as a combinatorial optimization task and employ a greedy algorithm to balance accuracy and efficiency, leading to our proposed method, GreedyPrune. Extensive experiments conducted across 9 datasets demonstrate that GreedyPrune consistently achieves state-of-the-art accuracy across various token pruning ratios, validating its effectiveness across diverse scenarios. Furthermore, GreedyPrune is compatible with different model sizes and LVLM (Large Vision-Language Model) architectures. Our results indicate that this method not only reduces the end-to-end processing time of LVLM models but also significantly lowers memory consumption, thereby enhancing overall computational efficiency.

\bibliography{aaai25}
\section{6 Appendix}

\subsection{6.1 Additional Results}
\begin{table*}[ht!]
\centering
\caption{Performance comparison of our method with Dart on Qwen2-VL under three different token pruning ratios.}
\label{tab:result_with_qwen2-vl}
\resizebox{\textwidth}{!}{
    \begin{tabular}{c|c|cc|ccccccccc|c}
        \toprule[2pt]
        ~ & \multirow{2}{*}{Method} & \multirow{2}{*}{K} & \multirow{2}{*}{\makecell{Retained \\token ratio}}& MME & POPE & MMB & Ocrbench & TextVQA & OK-VQA & Nocaps & Flickr30K & GQA &  \multirow{2}{*}{Avg}. \\
        ~ & ~ & ~ & ~ & P-score & F1 & Acc & Acc & EM & EM & CIDEr & CIDEr & EM & ~\\ 
        \midrule
         \rule{0pt}{10.5pt}\multirow{5}{*}{\rotatebox{90}{Qwen2-VL-7B}} & \cellcolor{gray!30}Original & \cellcolor{gray!30}- & \cellcolor{gray!30}-&\cellcolor{gray!30} 1663 & \cellcolor{gray!30} 88.3 & \cellcolor{gray!30} 80.5 & \cellcolor{gray!30} 814 & \cellcolor{gray!30} 82.6 & \cellcolor{gray!30} 50.7 & \cellcolor{gray!30} 103.4 & \cellcolor{gray!30} 78.6 & \cellcolor{gray!30} 61.7 & \cellcolor{gray!30}100\%\\ 
        \cmidrule{2-14}        
        \rule{0pt}{10.5pt}~ & DART & 2 & 33.3\% & 1612 & 87.5 & 78 & 676 & 78.2 & \textbf{50.1} & 98.3 & 71.5 & 59.7 & 94.69\% \\ 
        \rule{0pt}{10.5pt}~ & Ours($\tau$=0.8)&1 & 33.3\% & \textbf{1635} & \textbf{87.9} & \textbf{79.1} & \textbf{761} & \textbf{80.3} & 49.4 & \textbf{100.3} & \textbf{75.5} & \textbf{61.4} & \textbf{97.43\%} \\
        \cmidrule{2-14}
        \rule{0pt}{10.5pt}~ & DART & 2 & 22.2\% & 1557 & 85.8 & 75.1 & 600 & 73.5 & \textbf{49.4} &93.8 & 71.5 & 58.4 & 91.17\% \\ 
        \rule{0pt}{10.5pt}~ & Ours($\tau$=0.7)&1 & 22.2\% & \textbf{1626} & \textbf{87.7} & \textbf{76.9} & \textbf{667} & \textbf{76.4} & 49.3 & \textbf{99} & \textbf{74.5} & \textbf{61.1} & \textbf{94.87\%} \\
        \cmidrule{2-14}
        \rule{0pt}{10.5pt}~ & DART & 2 & 11.1\% & 1478 & 82.3 & 69.6 & 492 & 59.6 & 46.6 &84.4 & 63.7 & 55.3 & 82.82\% \\ 
        \rule{0pt}{10.5pt}~ & Ours($\tau$=0.6)&1 & 11.1\% & \textbf{1579} & \textbf{86.4} & \textbf{72.8} & \textbf{529} & \textbf{69.2} & \textbf{48} & \textbf{95.5} & \textbf{70.3} & \textbf{59.9} & \textbf{89.51\%} \\
        \bottomrule[2pt]
    \end{tabular}
}\label{table1}
\end{table*}
\begin{table*}[ht!]
\centering
\caption{The performance comparison between our method and other approaches on LLaVA-1.5-7B and LLaVA-1.5-13B under three different equivalent TFLOPS ratios.}
\label{tab:result_with_tflops}
\resizebox{\textwidth}{!}{
    \begin{tabular}{c|cc|cc|ccccccccc|c}
        \toprule[2pt]
        \multirow{2}{*}{Model} & \multirow{2}{*}{Method} & \multirow{2}{*}{TFLOPS} & \multirow{2}{*}{K} & \multirow{2}{*}{\makecell{Retained \\tokens}} & MME & POPE & MMB & Ocrbench & TextVQA & OK-VQA & Nocaps & Flickr30K & GQA & \multirow{2}{*}{Avg}. \\ 
        ~ & ~ & ~ & ~ & ~ & P-score & F1 & Acc & Acc & EM & EM & CIDEr & CIDEr & EM & ~\\ 
        \midrule
         \rule{0pt}{10.5pt}\multirow{13}{*}{\rotatebox{90}{LLaVA-1.5-7B} }& \cellcolor{gray!30}Original & \cellcolor{gray!30}100\% & \cellcolor{gray!30}- & \cellcolor{gray!30}- & \cellcolor{gray!30}1509 & \cellcolor{gray!30}85.8 & \cellcolor{gray!30}64 & \cellcolor{gray!30}314 & \cellcolor{gray!30}46 & \cellcolor{gray!30}53.4 & \cellcolor{gray!30}105.5 & \cellcolor{gray!30}74.8 & \cellcolor{gray!30}61.9 &\cellcolor{gray!30}100\% \\ 
        \cmidrule{2-15}
        \rule{0pt}{10.5pt}~ & FastV & 39.38\% & 3 & 131    & 1483             & 79              & \underline{63.2}& 292            & 43.4                             & 51.2           & \underline{103}& 72.5          & 58 & 95.62\% \\ 
        \rule{0pt}{10.5pt}~ & DART & 39.24\% & 2 & 179     & \textbf{1504}    & 82.3            & 63.1            & \textbf{301}   & \underline{43.9}                 & 51.3           & \textbf{103.6}   & \textbf{74.2} & 59.9 & 97.30\% \\ 
        \rule{0pt}{10.5pt}~ & DivPrune& 39.35\% & 0 & 205 & 1438             & \textbf{86.7}   & 62.9            & \underline{299}& 43.2                             & \textbf{52.1}  & 101.2            & 71.7          & \underline{60} & 96.67\% \\ 
        \rule{0pt}{10.5pt}~ & Ours($\tau$=0.94) & 39.23\% & 1 & 192     & \underline{1488} & \underline{85.5}& \textbf{63.3}   & \textbf{301}   & \textbf{44.4}                    &\underline{51.9}& 102.1            & \underline{73}& \textbf{61.4} & \textbf{97.81\%} \\
        \cmidrule{2-15}
        \rule{0pt}{10.5pt}~ & FastV & 29.42\% & 3 & 97 & 1403 & 73.6 & 62 & 275 & 40.1 & 49.1 & 98 & 68.2 & 55.7 &  90.71\% \\ 
        \rule{0pt}{10.5pt}~ & DART & 29.37\% & 2 & 113 & \underline{1477} & 79 & \textbf{62.6} & 282 & 42.1 & \underline{50.4} & \textbf{101.9} & \textbf{73.1} & 58.3 & 94.66\% \\
        \rule{0pt}{10.5pt}~ & DivPrune& 29.43\% & 0 & 143 & 1396 & \textbf{86.9} & \underline{62.5} & \textbf{293} & \underline{42.3} & 50.2 & 98.9 & 69.7 & \underline{59.6} & 94.88\% \\ 
        \rule{0pt}{10.5pt}~ & Ours($\tau$=0.86) & 29.32\% & 1 & 128 & \textbf{1483} & \underline{85.6} & 62.4 & \underline{291} & \textbf{43.6} & \textbf{51.2} & \underline{101.8} & \underline{72.2} & \textbf{61.2} & \textbf{96.75\%} \\ 
        \cmidrule{2-15}
        \rule{0pt}{10.5pt}~ & FastV & 19.58\% & 3 & 28 & 1064 & 49 & 51.3 & 169 & 28 & 38.6 & 66.4 & 41.4 & 48.6 & 65.73\% \\ 
        \rule{0pt}{10.5pt}~ & DART & 19.53\% & 2 & 46 & 1350 & 69.4 & 59.1 & 240 & 34.9 & 46.2 & 93 & 64.6 & 54 & 84.81\% \\ 
        \rule{0pt}{10.5pt}~ & DivPrune & 19.58\% & 0 & 81 & \underline{1370} & \textbf{86.8}  & \underline{60.1}  & \underline{281} & \underline{40.2}  & \underline{49.4}  & \underline{95.9}  & \underline{66.3}  & \underline{58.4}  & 92.13\% \\ 
        \rule{0pt}{10.5pt}~ & Ours($\tau$=0.78) & 19.49\% & 1 & 64 & \textbf{1427} & \underline{84.5}  & \textbf{61.0}  & \textbf{284}  & \textbf{41.9}  & \textbf{49.6}  & \textbf{98.4}  & \textbf{69.7}  & \textbf{60.2}  & \textbf{94.22\%} \\ 
        \midrule
         \multirow{5}{*}{\rotatebox{90}{LLaVA-1.5-13B}} & \cellcolor{gray!30}Original & \cellcolor{gray!30}100\% & \cellcolor{gray!30}- & \cellcolor{gray!30}-& \cellcolor{gray!30}1530 & \cellcolor{gray!30}85.9 & \cellcolor{gray!30}68.6 & \cellcolor{gray!30}338 & \cellcolor{gray!30}48.7 & \cellcolor{gray!30}58.2 & \cellcolor{gray!30}109.3 & \cellcolor{gray!30}79.4 & \cellcolor{gray!30}63.2 & \cellcolor{gray!30}100\%\\
        \cmidrule{2-15}
        \rule{0pt}{10.5pt} ~ & FastV & 19.54\% & 3 & 39 & 1295 & 61.2 & 59.8 & 178 & 30.1 & 46.9 & 84.7 & 56.3 & 53.2 & 72.01\% \\ 
        \rule{0pt}{10.5pt} ~ & DART & 19.62\% & 2 & 54 & 1389 & 71.8 & 64.3 & 243 & 33.3 & 51 & 99.3 & 71.6 & 55.7 & 82.49\% \\ 
        \rule{0pt}{10.5pt} ~ & DivPrune & 19.68\% & 0 & 81 & \underline{1463} & \textbf{85.1} & \underline{65.3} & \underline{294} & \underline{40.9} & \underline{55.2} & \underline{98.1} & \underline{69.9} & \underline{57.8} & 89.81\% \\ 
        \rule{0pt}{10.5pt} ~ & Ours($\tau$=0.78)&19.05\% & 1 & 64 & \textbf{1524} & \underline{85} & \textbf{65.5} & \textbf{323} & \textbf{45.3} & \textbf{55.5} & \textbf{102.4} & \textbf{74.5} & \textbf{61.6} & \textbf{94.98\%} \\
        \bottomrule[2pt]
    \end{tabular}
    }\label{table1}
\end{table*}
\subsubsection{Experiment on advanced LVLM}
To validate the extensibility of the proposed method to other advanced LVLMs, this study conducts image understanding experiments on the Qwen2-VL model. The benchmark pipeline adopts the official implemented DART and FastV algorithms on Qwen2-VL model as comparative baselines. Since FastV experienced OOM in multiple datasets and DART showed better performance than FastV, the specific data of FastV are no longer listed in the results. The experimental results are shown in Table \ref{tab:result_with_qwen2-vl}: under the specified visual token pruning conditions, the method proposed in this paper demonstrates better performance indicators than the baseline methods, providing empirical support for the effectiveness of this method in the application scenarios of the Qwen2-VL model.
\subsubsection{Experiment with aligned TFLOPS}
To enhance the depth of our performance-latency trade-off experiment, we transform the pruning strategies of various methods into equivalent TFLOPS. This enables a fair comparison of different algorithms under the same computational constraints. Additionally, we conduct extensive comparative experiments on the LLaVA-v1.5-7B model, evaluating various baseline methods across different equivalent TFLOPS conditions to gain deeper insights into their effectiveness.
Referencing the definition approach of FastV, this paper defines the TFLOPS ratio as the ratio of the TFLOPS of the model with visual tokens pruned to the TFLOPS of the original unpruned model. The calculation formula is as follows:
\begin{equation}
\resizebox{0.9\columnwidth}{!}{
$\frac{K \times (4\mu d^2 - 2\mu^2 d + 2\mu d m) + (T - K) \times (4\tilde{\mu} d^2 - 2\tilde{\mu}^2 d + 2\tilde{\mu} d m)}{T \times (4\mu d^2 - 2\mu^2 d + 2\mu d m)}$
}
\label{eq:relaxed_subproblem}
\end{equation}
$T$ denotes the total number of layers in the LLM component of the LVLM. $K$ represents the layer number where token pruning is applied. The sequence length before pruning is $\mu = N + M$, while the pruned sequence length is $\tilde{\mu} = N + \tilde{M}$, where $N$ represents the text sequence length. The parameter $d$ defines the hidden state dimension at each layer, and $m$ specifies the intermediate size of the feed-forward network module.
To ensure fairness, we standardized the TFLOPs across all baselines and the greedy pruning method in our experiments. Based on this standardization, we adjusted the final number of retained visual tokens while adhering to each baseline’s default parameters.

Table \ref{tab:result_with_tflops} presents a comparative analysis of the LLaVA series models against baseline methods under the different aligned TFLOPS ratios. Extensive experiments on LLaVA-1.5-7B and LLaVA-1.5-13B demonstrate that our method achieves outstanding results across various equivalent TFLOPS conditions. This validates its high practicality in both memory-sensitive and latency-sensitive scenarios.

\subsection{6.2 Datasets}
\subsubsection{Nocaps}Nocaps contains 166,100 manually generated descriptions to characterize 15,100 images from the validation and test sets of Open Images.
\subsubsection{Flickr30K}The Flickr30K dataset comprises over 30,000 images, each annotated with five descriptive sentences by humans.

\subsubsection{OK-VQA} OK-VQA includes 14,055 open-ended questions, with five ground-truth answers provided for each question.
\subsubsection{TextVQA} TextVQA consists of 45,336 questions across 28,408 images, all of which require textual reasoning to answer.
\subsubsection{GQA} GQA contains 22 million questions about various daily images, and each image is associated with a new streamlined scene graph based on Visual Genome.
\subsubsection{MME} MME dataset includes 14 subtasks, spanning multiple aspects from coarse-grained to fine-grained object recognition, commonsense reasoning, numerical calculation, text translation, and code reasoning.
\subsubsection{POPE} POPE transforms hallucination evaluation into a series of true/false questions about whether objects exist in images for the model to answer.
\subsubsection{MMBench} MMBench covers 20 fine-grained competencies, with approximately 3,000 multiple-choice questions collected from the internet and authoritative benchmark datasets.
\subsubsection{OCRbench} OCRbench was used as the benchmark. OCRBench comprises a collection of 1,000 manually screened and corrected question-answer pairs, covering five representative text-related tasks.\\
The system prompts are all default prompts provided in the lmms - evals evaluation package.

\subsection{6.3 Visualization of Retained Tokens}
\begin{figure*}[htbp]
  \centering
    \begin{subfigure}{0.49755\textwidth}
    \includegraphics[width=\linewidth]{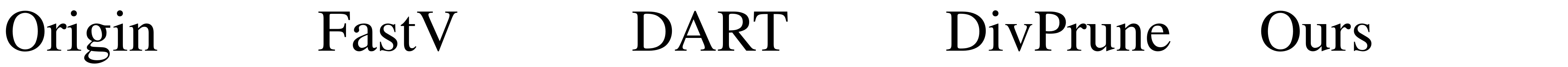}
  \end{subfigure}
  \hfill
  \begin{subfigure}{0.49755\textwidth}
    \includegraphics[width=\linewidth]{figs/picture_demo_14.pdf}
  \end{subfigure}
  \begin{subfigure}{0.49755\textwidth}
    \includegraphics[width=\linewidth]{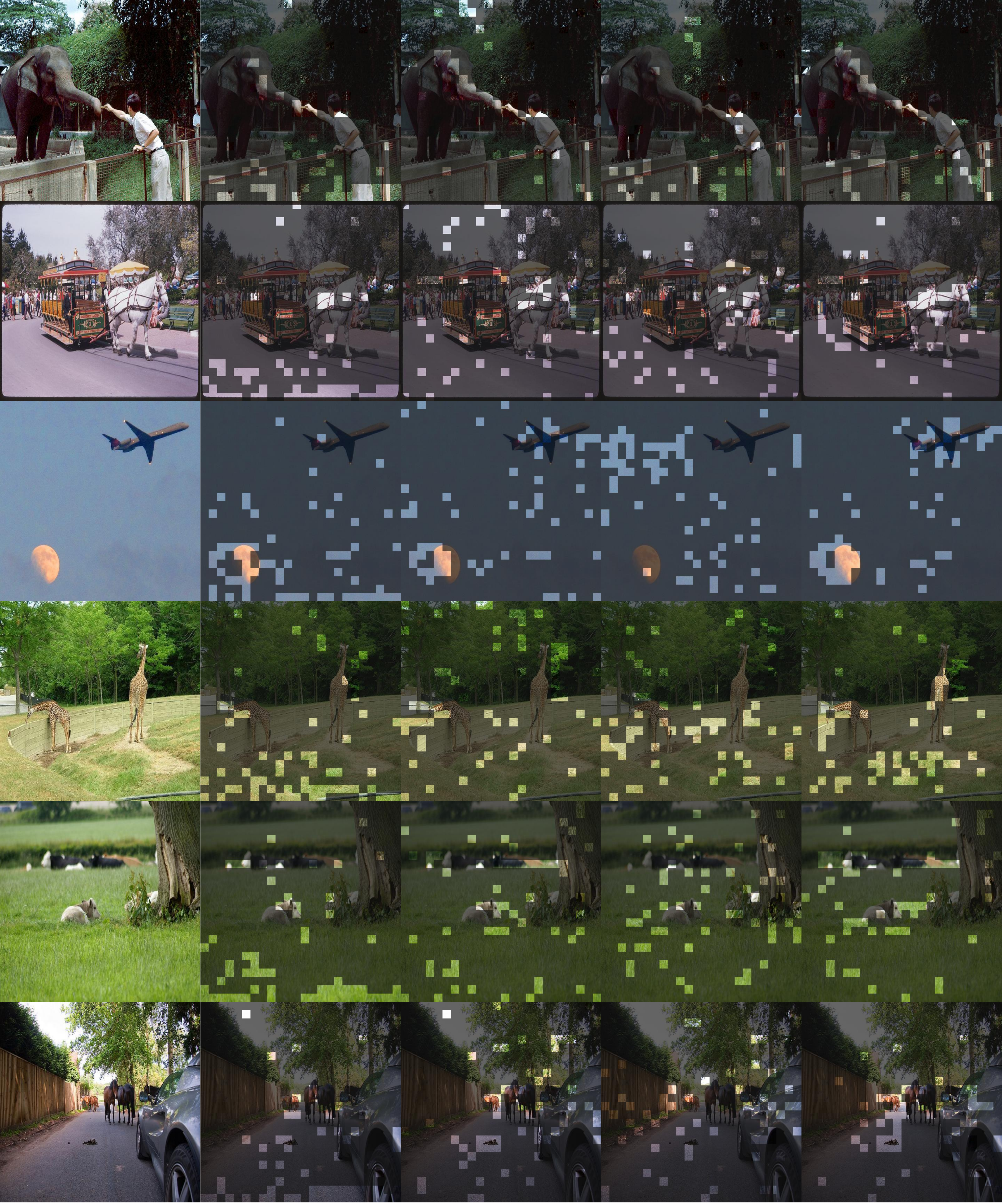}
  \end{subfigure}
  \hfill
  \begin{subfigure}{0.49755\textwidth}
    \includegraphics[width=\linewidth]{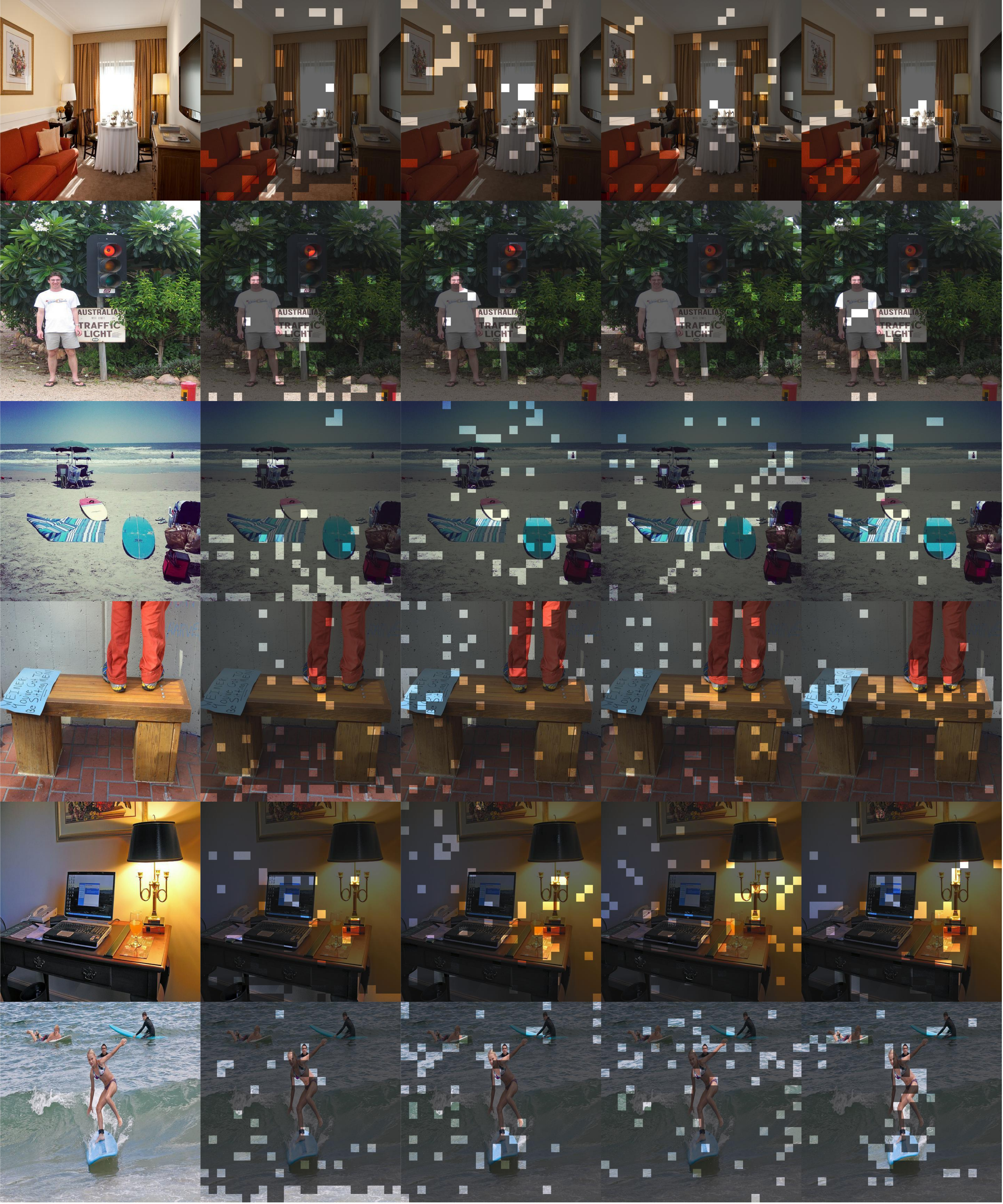}
  \end{subfigure}
  
  
  \begin{subfigure}{0.49755\textwidth}
    \includegraphics[width=\linewidth]{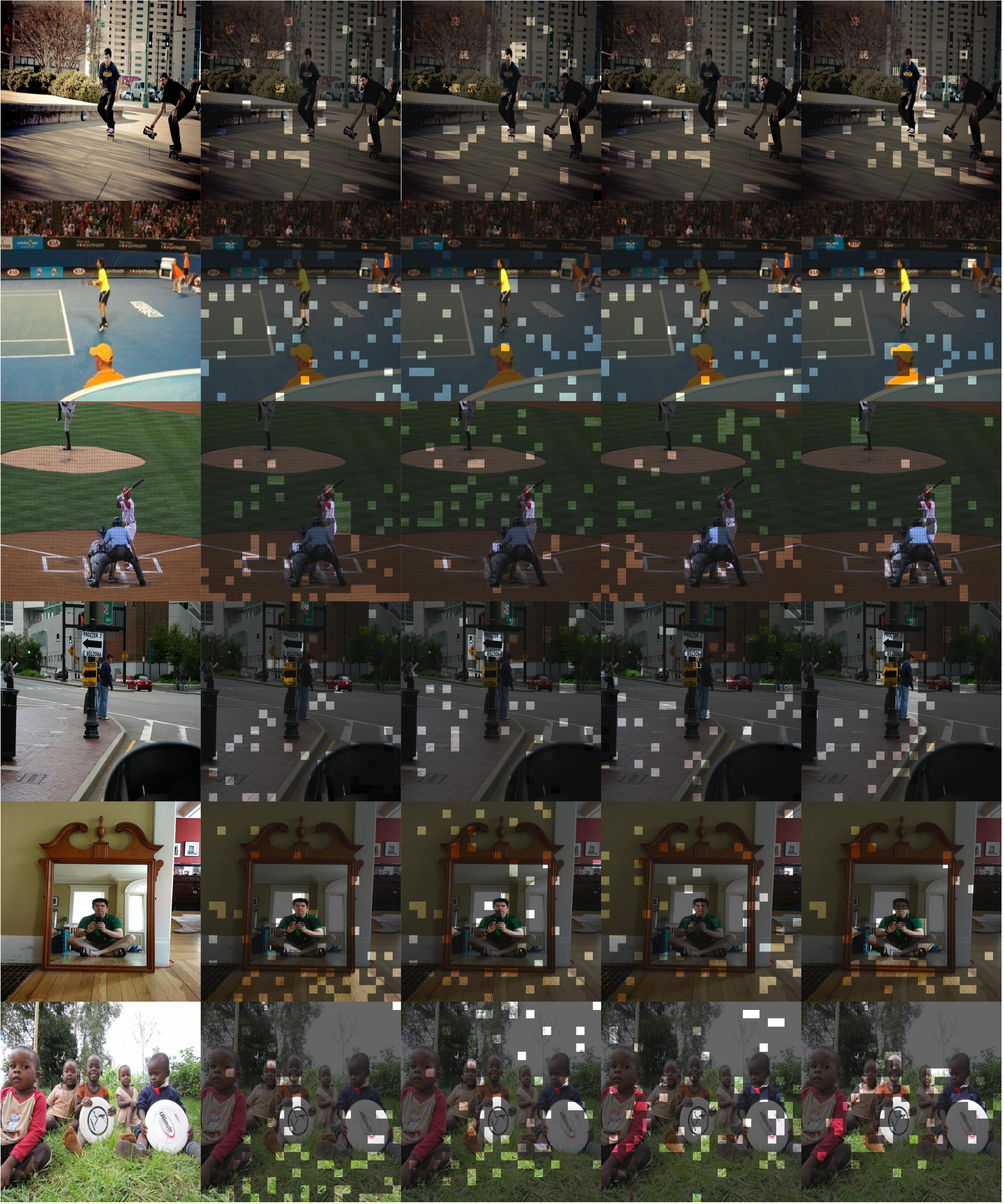}
  \end{subfigure}
  \hfill
  \begin{subfigure}{0.49755\textwidth}
    \includegraphics[width=\linewidth]{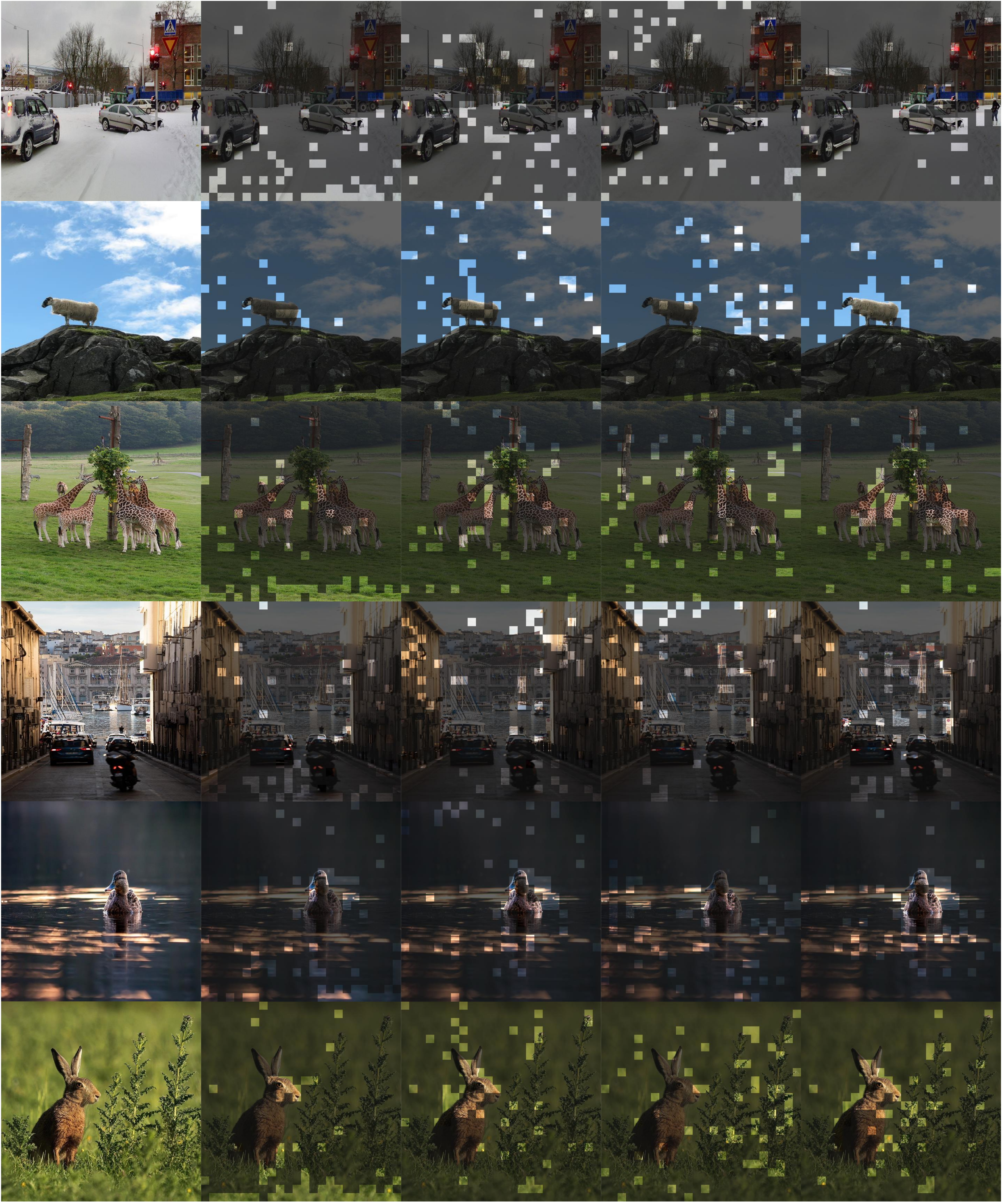}
  \end{subfigure}
  
  \caption{The columns in the figure are arranged from left to right as follows: input image, FastV, DART, Divprune, and the processing results of the proposed method under the cropping rate of 88.9\%}
  \label{fig:four_figures}
\end{figure*}

To intuitively and explainably demonstrate the effectiveness of our method, we visualized the positions of retained image tokens under an 88.9\% pruning ratio on LLaVA-1.5-7B. Figure ~\ref{fig:four_figures} presents visualizations of sample images and their retained visual tokens processed by each method. For each method, experimental validation was conducted using an 88.9\% pruning ratio. The visualization results show that the method proposed in this study demonstrates significant advantages in key information retention compared with other comparative methods.

\end{document}